\documentclass[10pt,a4paper]{article}
\usepackage[margin=1in]{geometry}
\usepackage[T1]{fontenc}
\usepackage[utf8]{inputenc}
\usepackage{lmodern}
\usepackage{microtype}
\usepackage{setspace}
\usepackage{amsmath}
\usepackage{amssymb}
\usepackage{booktabs}
\usepackage{array}
\usepackage{graphicx}
\usepackage{xcolor}
\usepackage{tikz}
\usepackage{authblk}
\usepackage[hidelinks]{hyperref}
\usepackage[numbers,sort&compress]{natbib}

\usetikzlibrary{arrows.meta,positioning,fit,calc}
\newcommand{\RowNet}{\textsc{RowNet}}
\newcommand{\R}{\mathbb{R}}
\newcommand{\eps}{\varepsilon}
\newcommand{\softmax}{\operatorname{softmax}}
\newcommand{\MAPE}{\operatorname{MAPE}}

\title{RowNet: A Memory Transformer for Tabular Regression}

\author{
Askat Rakhymbekov \\
Department of Applied Mathematics and Informatics \\
Kyrgyz-Turkish Manas University \\
Bishkek, Kyrgyzstan \\
\texttt{2312.01002@manas.edu.kg}
\and
Gulshat Muhametjanova \\
Department of Applied Mathematics and Informatics \\
Kyrgyz-Turkish Manas University \\
Bishkek, Kyrgyzstan \\
\texttt{gulshat.muhametjanova@manas.edu.kg}
}

\date{}

\begin{document}
\maketitle

\begin{abstract}
Real estate valuation is a structured regression problem in which prices are governed by heterogeneous feature types, sparse regional effects, nonlinear interactions, and the practical logic of comparable properties. Standard multilayer perceptrons treat each row as an isolated vector and must learn locality, scale sensitivity, and categorical matching from supervision alone. Gradient-boosted decision trees provide strong tabular baselines, but their feature-centric splitting mechanism does not explicitly model the retrieval of similar historical observations. This paper presents \RowNet, a retrieval-based neural architecture for real estate price-per-square-meter prediction. \RowNet{} represents a query property through pairwise similarity features against a memory bank of labeled properties. A first retrieval layer estimates a coarse target from feature-only similarities. A second layer augments the memory comparison with target-consistency features and uses multiple learned attention heads to retrieve complementary comparable sets. A final mixture-of-experts module combines learned gating, residual correction, entropy regularization, and head-diversity regularization to produce the prediction. The model is intentionally compact: it uses exact categorical matches, three scale-aware numerical similarity kernels, softmax row attention over the memory bank, multi-head linear compatibility scoring, and per-head residual networks. Experiments on a real-world real estate valuation dataset from Bishkek, Kyrgyzstan demonstrate the effectiveness of the proposed approach. \RowNet{} achieved a competition MAPE of 7.44\%, demonstrating that retrieval-based neural architectures can effectively exploit comparable-property relationships for real estate valuation. The study argues that learned row retrieval is a natural inductive bias for real estate valuation: it combines neural metric learning with the domain practice of estimating a property from comparable transactions.
\end{abstract}

\noindent\textbf{Keywords:} tabular regression, real estate valuation, retrieval-augmented learning, row attention, memory networks, mixture of experts, comparable properties.

\section{Introduction}

Tabular regression remains one of the most difficult regimes for deep learning. Unlike images or text, tabular data do not provide a stable local topology: adjacent columns are rarely semantically adjacent, categorical identifiers can have thousands of sparse levels, numerical features live on incompatible scales, and missingness often carries domain-specific meaning. In real estate valuation these difficulties become sharper. A property's observed price depends on structural factors such as area, floor, building age, ceiling height, and number of rooms; geographic factors such as district, street, coordinate cluster, and distance from the urban center; listing dynamics such as recency, views, and user interest; and qualitative attributes such as building series, heating type, renovation status, furniture, parking, security, legal documents, and utilities. The mapping from these features to price is neither globally smooth nor purely additive.

The key difficulty is that real estate markets are locally heterogeneous. A square meter in one district is not comparable to a square meter in another district without considering building type, floor, condition, accessibility, and neighborhood supply. Even within the same city, price functions change across micro-regions and property submarkets. A global parametric model must therefore learn many conditional regimes. A tree ensemble can capture regime changes through splits, but it still operates as a feature-centric partition of the input space. A multilayer perceptron can learn nonlinear interactions, but it receives a single row and must infer from its parameters which historical examples are relevant. Neither paradigm directly expresses the procedure used by human appraisers: identify comparable properties, weight them according to similarity, and adjust the estimate by systematic residual factors.

We adopt a comparable-property inductive bias and implement a retrieval-based model specialized for real estate valuation. The architecture abandons a generic feature-token transformer in favor of a learned row-retrieval model over engineered pairwise similarities. The model predicts price per square meter, then multiplies by area to recover the submitted total price.

\RowNet{} differs from a conventional MLP in three ways. First, it does not treat a training row only as an independent input-output pair. During both training and inference, the full training set is a memory bank. A query listing retrieves from this memory through attention weights over rows. Second, the attention score is not a raw dot product in the original feature space. It is computed from a pairwise similarity vector containing exact categorical matches and scale-aware numerical similarity functions. Third, prediction is not a single attention average. \RowNet{} performs a coarse feature-only retrieval, then refines retrieval by adding target-consistency features derived from the first-stage estimate, and finally combines multiple retrieval heads through a mixture-of-experts gate and a residual correction.

The resulting architecture is intentionally hybrid. It keeps the statistical strengths of nearest-neighbor estimation, the differentiability of attention mechanisms, the specialization of multi-head retrieval, and the calibration flexibility of residual prediction. Its inductive bias is particularly suitable for real estate markets: prices are often easiest to estimate by asking which historical properties are comparable, but comparability itself must be learned from data.

\subsection{Contributions}

This paper makes the following contributions.

\begin{enumerate}
    \item We present \RowNet, a compact retrieval-augmented tabular regression architecture. The method uses a memory bank of labeled rows, pairwise similarity embeddings, two-stage row attention, multi-head retrieval, mixture-of-experts aggregation, and residual correction.
    \item We formalize the architecture mathematically, including categorical match features, numerical similarity kernels, target-conditioned retrieval features, head-wise softmax attention, gating, softmin residual weighting, entropy regularization, and diversity regularization.
    \item We describe the preprocessing pipeline: Russian-to-English schema normalization, missing-value handling, geographic clustering, distance-to-center features, address decomposition, listing-time parsing, area transformations, address statistics, multi-label feature expansion, one-hot encoding, label encoding, and price-per-square-meter target construction.
    \item We report the training trace, a proof-of-concept retrieval baseline, and the competition score, and discuss qualitative comparisons where numeric evidence is unavailable.
    \item We analyze \RowNet{} as a learned comparable-property model and discuss its relation to gradient boosting, tabular transformers, memory networks, retrieval-based learning, and $k$-nearest-neighbor methods.
\end{enumerate}

\section{Related Work}

\subsection{Gradient-Boosted Trees for Tabular Data}

Gradient-boosted decision trees remain the dominant baseline for many tabular tasks. XGBoost \citep{chen2016xgboost} popularized scalable second-order boosting with regularized tree learners and remains a strong default for structured prediction. LightGBM \citep{ke2017lightgbm} improves training efficiency through histogram-based splitting, leaf-wise growth, and gradient-based one-side sampling. CatBoost \citep{prokhorenkova2018catboost} directly addresses categorical variables through ordered target statistics and combats target leakage in categorical encoding. These methods are powerful because decision trees naturally handle heterogeneous feature scales, nonlinear thresholds, sparse interactions, and missing values.

However, boosted trees are feature-centric. A prediction is obtained by routing a row through many learned partitions and summing leaf values. The model can approximate local neighborhoods implicitly, but it does not explicitly retrieve comparable historical rows. In real estate valuation this distinction matters. A boosted tree may learn that a certain area and district combination is expensive, but it does not expose or optimize the set of comparable listings used for a particular prediction. \RowNet{} instead builds prediction around a learned attention distribution over memory rows. This does not make boosting obsolete; rather, it offers an alternative inductive bias for markets where local comparability is central.

\subsection{Deep Learning for Tabular Data}

Deep tabular models attempt to adapt representation learning to heterogeneous structured features. TabNet \citep{arik2021tabnet} uses sequential attentive feature selection, allowing each decision step to focus on a sparse subset of features. TabTransformer \citep{huang2020tabtransformer} contextualizes categorical embeddings with transformer layers and is especially relevant when categorical variables are high-cardinality and semantically interacting. FT-Transformer \citep{gorishniy2021rtdl} tokenizes continuous and categorical features and applies transformer blocks to feature tokens. SAINT \citep{somepalli2021saint} combines attention over features and rows, introducing intersample attention and contrastive pretraining for tabular data.

\RowNet{} is related to these models through attention, but its attention axis and representation differ. FT-Transformer and TabTransformer primarily build feature representations inside a row. SAINT explicitly studies row attention, but \RowNet{} does not use a generic transformer encoder over row embeddings. Instead, it constructs pairwise query-memory similarity vectors and learns compatibility weights over those vectors. The value retrieved by attention is the target itself, not a learned hidden token. This makes \RowNet{} closer to differentiable nearest-neighbor regression than to feature-token transformers.

\subsection{Memory-Augmented and Retrieval-Based Learning}

Memory networks \citep{weston2014memory,sukhbaatar2015endtoend} store external facts or examples and retrieve them through learned attention. Retrieval-augmented models have become common in language and vision because they allow parametric models to condition on nonparametric evidence. The classical $k$-nearest-neighbor classifier \citep{cover1967nearest} can also be understood as a nonparametric memory method, where the memory consists of labeled training examples and inference is local averaging or voting.

Recent tabular retrieval approaches such as TabR \citep{gorishniy2023tabr} show that combining learned representations with nearest-neighbor retrieval can improve tabular prediction. \RowNet{} follows the same broad motivation, but it is specifically engineered around real estate comparables. Its memory key is not a single latent vector. Instead, every query-memory pair is transformed into a structured similarity vector containing exact matches, inverse distance, exponential relative similarity, clipped linear relative similarity, and target-consistency features. The model then learns how to combine these similarity channels through attention heads and a mixture-of-experts aggregator.

\subsection{Attention Mechanisms and Residual Design}

The transformer attention mechanism \citep{vaswani2017attention} computes compatibility between queries and keys, normalizes scores with a softmax, and aggregates values. \RowNet{} uses the same normalized weighted-sum principle, but with a task-specific compatibility function. Rather than computing $QK^\top/\sqrt{d}$ from learned projections, it computes a dot product between learned head weights and pairwise similarity features. This choice emphasizes interpretable similarity channels and stable computation over raw feature-space attention.

We also relate the design to depth-wise attention over residual layers \citep{kimi2026attentionresiduals}. \RowNet{} instead uses residual prediction in the sense of an additive correction to the retrieved target average, not transformer residual blocks with LayerNorm.

\section{Methodology}

\subsection{Problem Definition}

Let the training set be
\begin{equation}
    \mathcal{D} = \{(\mathbf{c}_i, \mathbf{x}_i, y_i)\}_{i=1}^{N},
\end{equation}
where $\mathbf{c}_i \in \R^{p}$ denotes categorical or binary-coded features, $\mathbf{x}_i \in \R^{q}$ denotes numerical features, and $y_i \in \R_{+}$ is the target price per square meter. In our pipeline, $y_i$ is constructed as
\begin{equation}
    y_i = \frac{\texttt{usd\_price}_i}{\texttt{area}_i}.
\end{equation}
At inference time the predicted total property price is recovered by
\begin{equation}
    \widehat{\texttt{usd\_price}}_i = \widehat{y}_i \cdot \texttt{area}_i.
\end{equation}
The decision to predict price per square meter rather than total price is important: it removes a dominant linear area effect and asks the model to learn the market value density conditioned on location, building attributes, and amenities.

\subsection{Preprocessing and Feature Engineering}

The preprocessing pipeline maps raw columns from Russian labels to English feature names. Known low-utility or sparsely useful infrastructure columns are removed, including land-area and utility-related fields, and the test identifier is excluded from modeling. Missing categorical values are filled with \texttt{NA}; missing numerical values are filled with medians; and remaining incomplete training rows are removed as a final safety step.

Temporal listing fields are parsed into continuous day counts. For example, strings equivalent to ``added seven days ago'' or ``raised eight hours ago'' are mapped into \texttt{added\_days} and \texttt{upped\_days}. The address field is decomposed into two address components, \texttt{address\_1} and \texttt{address\_2}, because different address granularities carry different market signals. Geographic coordinates are processed jointly over train and test rows: valid latitude-longitude pairs are clustered with $K=20$ KMeans clusters, the median coordinate is used as the empirical city center, and a haversine distance-to-center feature is computed. Latitude and longitude are then removed from the final feature table.

The building descriptor is decomposed into material and construction year. The main property descriptor is parsed into number of rooms and area; from these the pipeline derives \texttt{area\_per\_room} and \texttt{room\_per\_area}. Floor information is parsed into current floor, maximum floor, and \texttt{floor\_ratio}. Construction year is transformed into building age using a reference year of 2025. Ceiling height is converted to a numeric value, with missing or zero values replaced by a mode-based default.

Several multi-label textual fields are expanded into binary indicators. Legal documents become \texttt{doc\_*} columns, security attributes become \texttt{sec\_*} columns, and miscellaneous amenities become \texttt{misc\_*} columns. The original multi-label fields are also converted into counts. Area-based nonlinear transformations are added:
\begin{equation}
    \log(1+\texttt{area}), \qquad \sqrt{\max(\texttt{area},0)}, \qquad \texttt{area}^2, \qquad \texttt{area}\cdot \texttt{ceiling\_height}.
\end{equation}
Finally, address statistics are computed for both address levels: counts, average age, average maximum floor, average view count, average hearts, average added days, and average raised days. These are unsupervised or non-target descriptive statistics in the final label-encoding pipeline used by \RowNet.

The final preprocessing uses one-hot encoding for 16 non-address categorical fields, producing 89 one-hot columns, and label encoding for \texttt{address\_1}, \texttt{address\_2}, and \texttt{geo\_cluster}. The processed training table contains 7,128 rows and 156 columns; the processed test table contains 1,784 rows and 156 columns. The \RowNet{} implementation then separates columns into a categorical set and a numerical set. Importantly, many binary one-hot and multi-label indicators are deliberately treated as categorical for similarity computation: exact equality on such columns is a meaningful signal that two properties share the same discrete attribute.

\subsection{Architecture Overview}

\RowNet{} is implemented as \texttt{ThreeLayerMoE}. The name reflects three computational stages:
\begin{enumerate}
    \item feature-only row retrieval,
    \item target-conditioned multi-head row retrieval,
    \item mixture-of-experts aggregation with residual correction.
\end{enumerate}

The model parameters are small relative to conventional neural networks. The first retrieval layer has a single learnable vector $\mathbf{w}_1 \in \R^{d_1}$. The second retrieval layer has $H$ learnable head vectors $\mathbf{w}_{2,h} \in \R^{d_2}$, where $H=8$. The mixture gate is a two-layer MLP
\begin{equation}
    G(\mathbf{z}) = W_g^{(2)} \sigma(W_g^{(1)}\mathbf{z}+\mathbf{b}_g^{(1)})+\mathbf{b}_g^{(2)},
\end{equation}
with hidden size 128 and ReLU activation. Each head has its own residual MLP
\begin{equation}
    R_h(\mathbf{z}) = W_{r,h}^{(2)} \sigma(W_{r,h}^{(1)}\mathbf{z}+\mathbf{b}_{r,h}^{(1)}) + b_{r,h}^{(2)},
\end{equation}
also with hidden size 128.

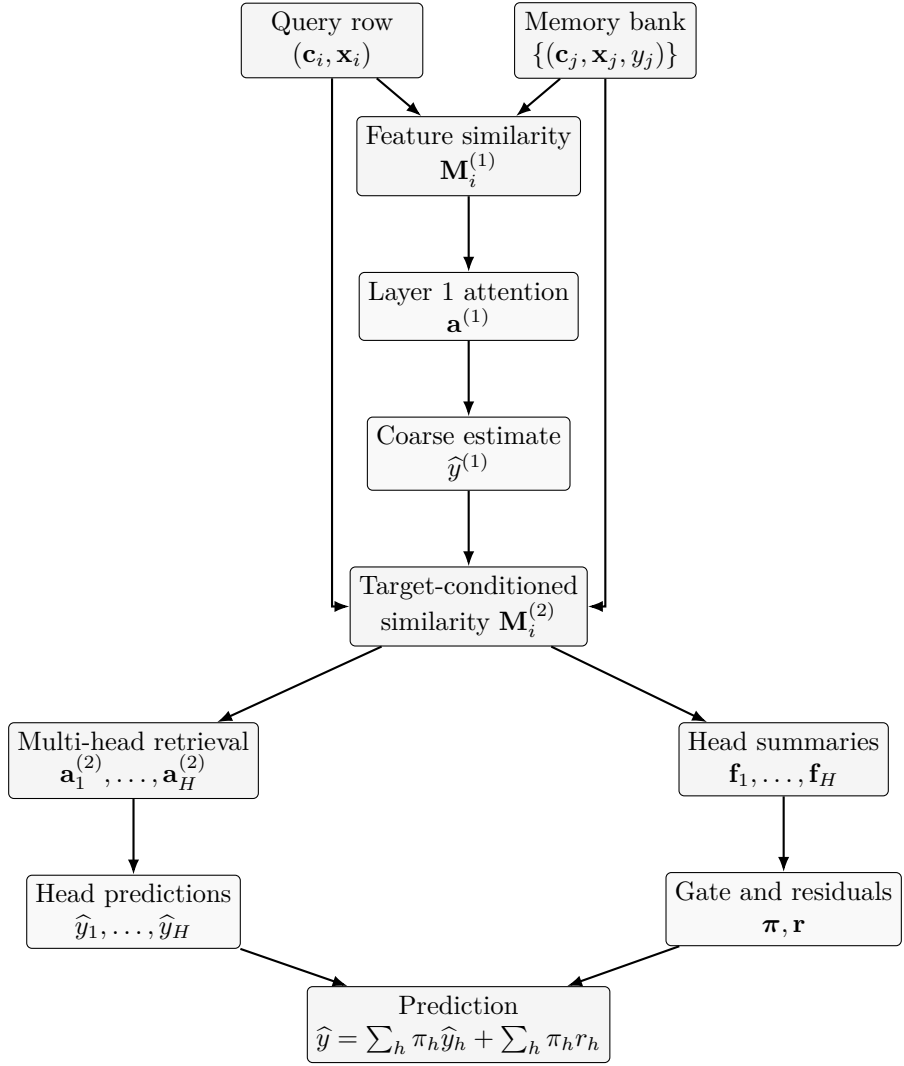
\begin{figure}[t]
\centering
\begin{tikzpicture}[
    node distance=1.0cm and 1.2cm,
    block/.style={draw, rounded corners=2pt, align=center, minimum height=0.8cm, minimum width=2.4cm, fill=gray!8},
    small/.style={draw, rounded corners=2pt, align=center, minimum height=0.65cm, minimum width=2.0cm, fill=gray!5},
    arrow/.style={-{Latex[length=2mm]}, thick}
]
\node[block] (q) {Query row\\$(\mathbf{c}_i,\mathbf{x}_i)$};
\node[block, right=of q] (mem) {Memory bank\\$\{(\mathbf{c}_j,\mathbf{x}_j,y_j)\}$};
\node[block, below=of $(q)!0.5!(mem)$] (m1) {Feature similarity\\$\mathbf{M}^{(1)}_i$};
\node[small, below=of m1] (a1) {Layer 1 attention\\$\mathbf{a}^{(1)}$};
\node[small, below=of a1] (y1) {Coarse estimate\\$\widehat{y}^{(1)}$};
\node[block, below=of y1] (m2) {Target-conditioned\\similarity $\mathbf{M}^{(2)}_i$};
\node[block, below left=of m2] (heads) {Multi-head retrieval\\$\mathbf{a}^{(2)}_1,\ldots,\mathbf{a}^{(2)}_H$};
\node[block, below right=of m2] (feat) {Head summaries\\$\mathbf{f}_1,\ldots,\mathbf{f}_H$};
\node[small, below=of heads] (yh) {Head predictions\\$\widehat{y}_1,\ldots,\widehat{y}_H$};
\node[small, below=of feat] (gate) {Gate and residuals\\$\boldsymbol{\pi},\mathbf{r}$};
\node[block, below=1.0cm of $(yh)!0.5!(gate)$] (out) {Prediction\\$\widehat{y}=\sum_h\pi_h\widehat{y}_h+\sum_h\pi_h r_h$};

\draw[arrow] (q) -- (m1);
\draw[arrow] (mem) -- (m1);
\draw[arrow] (m1) -- (a1);
\draw[arrow] (a1) -- (y1);
\draw[arrow] (y1) -- (m2);
\draw[arrow] (q) |- (m2);
\draw[arrow] (mem) |- (m2);
\draw[arrow] (m2) -- (heads);
\draw[arrow] (m2) -- (feat);
\draw[arrow] (heads) -- (yh);
\draw[arrow] (feat) -- (gate);
\draw[arrow] (yh) -- (out);
\draw[arrow] (gate) -- (out);
\end{tikzpicture}
\caption{Overview of \RowNet{}. The model is a transformer-inspired row-attention system, but its compatibility scores are computed from engineered query-memory similarity vectors rather than learned $QK^\top$ projections.}
\label{fig:architecture}
\end{figure}

\subsection{Tensor Flow}

For a single query $i$, the training memory contains all $N$ processed training rows. During training, the query row is masked out of its own memory by setting its score to a large negative value before the softmax. During inference, the query comes from the test set and the memory is the full training set, so no self-mask is needed.

Let $p$ be the number of categorical or exact-match columns and $q$ the number of numerical columns. The first similarity matrix has shape
\begin{equation}
    \mathbf{M}^{(1)}_i \in \R^{N \times d_1}, \qquad d_1=p+3q.
\end{equation}
The first attention layer maps it to scores $\mathbf{s}^{(1)}\in\R^N$, attention weights $\mathbf{a}^{(1)}\in\R^N$, and a scalar coarse prediction $\widehat{y}^{(1)}_i$.

The second similarity matrix augments the first matrix with two target-similarity features and has shape
\begin{equation}
    \mathbf{M}^{(2)}_i \in \R^{N \times d_2}, \qquad d_2=d_1+2.
\end{equation}
With $H$ heads, the second score tensor has shape $N \times H$. Softmax normalization is applied over the memory dimension independently for each head. This yields $H$ attention distributions, $H$ retrieved target estimates, and $H$ retrieved feature summaries. The gate and residual modules then operate over the head dimension.

\subsection{Why These Components Exist}

The first layer provides a stable, feature-only comparable-property estimate. It answers the question: ``which historical properties look similar to the query before considering price?'' This stage is analogous to a learned soft nearest-neighbor estimator. It is deliberately simple, because its output is used not as the final prediction but as a bootstrap signal for target-conditioned refinement.

The second layer adds two features measuring how close each memory target $y_j$ is to the first-stage estimate. This is a form of teacher-forced retrieval during training and self-consistent retrieval during inference. It narrows attention toward rows that are not only feature-similar but also plausible under the coarse market estimate. In real estate terms, it discourages the model from comparing a query to properties whose observed price density is incompatible with the first retrieval pass.

The multi-head design allows different compatibility functions to specialize. One head may place more emphasis on location matches, another on area and room geometry, another on amenities or legal-document indicators, and another on target consistency. The implementation does not force these interpretations, but the head-diversity penalty discourages all heads from collapsing to the same linear scorer.

The mixture gate learns how much to trust each head from the retrieved head summaries. The residual MLPs learn small additive corrections that compensate for systematic bias in retrieved averages. The final softmin weighting over residual magnitudes favors heads whose residual correction is small in absolute value, while the learned gate remains free to emphasize informative heads. The final weight vector is a convex combination of these two selection mechanisms.

\section{Mathematical Formulation}

\subsection{Pairwise Similarity Embedding}

For a query row $i$ and memory row $j$, categorical similarity is defined as exact equality:
\begin{equation}
    \phi^{\mathrm{cat}}_{ijk} = \mathbf{1}[c_{ik}=c_{jk}], \qquad k=1,\ldots,p.
\end{equation}
This includes true categorical identifiers and binary one-hot indicators. Treating one-hot indicators as exact-match categorical features is useful because equality distinguishes shared presence, shared absence, and mismatch in a simple way.

For each numerical feature $l=1,\ldots,q$, define the absolute difference
\begin{equation}
    \Delta_{ijl}=|x_{jl}-x_{il}|,
\end{equation}
and a query-relative scale
\begin{equation}
    \alpha_{il}=|x_{il}|+\eps,
\end{equation}
where $\eps=10^{-6}$ in the implementation. \RowNet{} uses three numerical similarity kernels:
\begin{align}
    \phi^{(1)}_{ijl} &= \frac{1}{1+\Delta_{ijl}}, \\
    \phi^{(2)}_{ijl} &= \exp\left(-\frac{\Delta_{ijl}}{\alpha_{il}}\right), \\
    \phi^{(3)}_{ijl} &= \max\left(0,1-\frac{\Delta_{ijl}}{\alpha_{il}}\right).
\end{align}
These kernels express complementary notions of closeness. The inverse-distance feature is stable near zero and decays slowly. The exponential feature is scale-aware and sharply penalizes large relative deviations. The clipped linear feature provides a bounded local similarity that becomes exactly zero once the relative difference exceeds one.

The first-layer pairwise similarity vector is
\begin{equation}
    \mathbf{m}^{(1)}_{ij}
    =
    \left[
    \boldsymbol{\phi}^{\mathrm{cat}}_{ij},
    \boldsymbol{\phi}^{(1)}_{ij},
    \boldsymbol{\phi}^{(2)}_{ij},
    \boldsymbol{\phi}^{(3)}_{ij}
    \right]
    \in \R^{d_1}.
\end{equation}
Stacking these vectors over all memory rows gives $\mathbf{M}^{(1)}_i$.

\subsection{Layer 1: Feature-Only Retrieval}

The first retrieval layer scores each memory row with a learned linear compatibility function:
\begin{equation}
    s^{(1)}_{ij} = \left(\mathbf{m}^{(1)}_{ij}\right)^\top \mathbf{w}_1.
\end{equation}
During training, self-attention is removed:
\begin{equation}
    s^{(1)}_{ii} = -10^9.
\end{equation}
The attention distribution is
\begin{equation}
    a^{(1)}_{ij}
    =
    \frac{\exp(s^{(1)}_{ij})}{\sum_{r=1}^{N}\exp(s^{(1)}_{ir})}.
\end{equation}
The coarse prediction is the attention-weighted memory target:
\begin{equation}
    \widehat{y}^{(1)}_i
    =
    \sum_{j=1}^{N}a^{(1)}_{ij}y_j.
\end{equation}
This is a differentiable analogue of a weighted comparable-property estimate.

\subsection{Layer 2: Target-Conditioned Similarity}

The second retrieval layer augments the pairwise feature vector with target consistency relative to $\widehat{y}^{(1)}_i$. Define
\begin{equation}
    \delta_{ij}=|y_j-\widehat{y}^{(1)}_i|, \qquad
    \beta_i=|\widehat{y}^{(1)}_i|+\eps.
\end{equation}
The implementation uses two target-similarity features:
\begin{align}
    \psi^{(1)}_{ij} &= \frac{1}{1+\delta_{ij}/\beta_i},\\
    \psi^{(2)}_{ij} &= \exp\left(-\frac{\delta_{ij}}{\beta_i}\right).
\end{align}
The second-layer vector is
\begin{equation}
    \mathbf{m}^{(2)}_{ij}
    =
    \left[
    \mathbf{m}^{(1)}_{ij},
    \psi^{(1)}_{ij},
    \psi^{(2)}_{ij}
    \right]
    \in \R^{d_2}.
\end{equation}

\subsection{Multi-Head Retrieval}

For each head $h\in\{1,\ldots,H\}$, \RowNet{} learns a separate vector $\mathbf{w}_{2,h}\in\R^{d_2}$ and computes
\begin{equation}
    s^{(2)}_{ijh} = \left(\mathbf{m}^{(2)}_{ij}\right)^\top \mathbf{w}_{2,h}.
\end{equation}
The training self-mask is again applied:
\begin{equation}
    s^{(2)}_{iih}=-10^9.
\end{equation}
Attention is normalized over memory rows for each head:
\begin{equation}
    a^{(2)}_{ijh}
    =
    \frac{\exp(s^{(2)}_{ijh})}{\sum_{r=1}^{N}\exp(s^{(2)}_{irh})}.
\end{equation}
The head prediction is
\begin{equation}
    \widehat{y}_{ih} = \sum_{j=1}^{N} a^{(2)}_{ijh}y_j,
\end{equation}
and the retrieved head summary is
\begin{equation}
    \mathbf{f}_{ih} = \sum_{j=1}^{N}a^{(2)}_{ijh}\mathbf{m}^{(2)}_{ij}.
\end{equation}
The summary $\mathbf{f}_{ih}$ is a differentiable description of what head $h$ retrieved for query $i$.

\subsection{Mixture Gate and Residual Correction}

The global retrieval summary is the mean of head summaries:
\begin{equation}
    \bar{\mathbf{f}}_i = \frac{1}{H}\sum_{h=1}^{H}\mathbf{f}_{ih}.
\end{equation}
The learned gate is
\begin{equation}
    \mathbf{g}_i = \softmax(G(\bar{\mathbf{f}}_i)) \in \R^{H}.
\end{equation}
Each residual expert predicts a scalar correction:
\begin{equation}
    r_{ih}=R_h(\mathbf{f}_{ih}).
\end{equation}
The model also computes a residual-confidence distribution by applying a softmin over absolute residual magnitudes:
\begin{equation}
    \mathbf{u}_i
    =
    \softmax\left(-|r_{i1}|,\ldots,-|r_{iH}|\right).
\end{equation}
The final mixture weights combine the learned gate and residual-confidence weights:
\begin{equation}
    \boldsymbol{\pi}_i
    =
    \frac{\gamma\mathbf{g}_i+(1-\gamma)\mathbf{u}_i}
    {\mathbf{1}^\top\left(\gamma\mathbf{g}_i+(1-\gamma)\mathbf{u}_i\right)+10^{-8}},
\end{equation}
where $\gamma=0.5$ in the training run and in the default prediction function.

The final prediction is
\begin{equation}
    \widehat{y}_i
    =
    \underbrace{\sum_{h=1}^{H}\pi_{ih}\widehat{y}_{ih}}_{\text{retrieved price density}}
    +
    \underbrace{\sum_{h=1}^{H}\pi_{ih}r_{ih}}_{\text{residual correction}}.
\end{equation}
The residual may be positive or negative. This is necessary because a retrieved average can underprice or overprice the query depending on unobserved quality, local demand, or systematic feature bias.

\subsection{Objective Function}

The primary loss is mean absolute percentage error on price per square meter:
\begin{equation}
    \mathcal{L}_{\mathrm{MAPE},i}
    =
    \left|\frac{y_i-\widehat{y}_i}{y_i+\eps}\right|.
\end{equation}
The residual penalty controls the scale of the additive correction:
\begin{equation}
    \mathcal{L}_{\mathrm{res},i}
    =
    \left(\frac{\sum_h \pi_{ih}r_{ih}}{y_i+\eps}\right)^2.
\end{equation}
The gate entropy is
\begin{equation}
    \mathcal{H}(\mathbf{g}_i)
    =
    -\sum_{h=1}^{H}g_{ih}\log(g_{ih}+10^{-8}).
\end{equation}
Because the implementation subtracts this term from the loss, it encourages non-collapsed gate usage:
\begin{equation}
    -\lambda_{\mathrm{ent}}\mathcal{H}(\mathbf{g}_i).
\end{equation}
Head diversity is regularized through the normalized second-layer head weights. Let
\begin{equation}
    \widetilde{\mathbf{w}}_{2,h}
    =
    \frac{\mathbf{w}_{2,h}}{\|\mathbf{w}_{2,h}\|_2+10^{-8}},
\end{equation}
and let $\widetilde{\mathbf{W}}_2$ stack these row-wise. The diversity penalty is
\begin{equation}
    \mathcal{L}_{\mathrm{div}}
    =
    \frac{1}{H^2}
    \left\|
    \widetilde{\mathbf{W}}_2\widetilde{\mathbf{W}}_2^\top - \mathbf{I}_H
    \right\|_F^2.
\end{equation}
The per-query training objective is
\begin{equation}
    \mathcal{L}_i
    =
    \mathcal{L}_{\mathrm{MAPE},i}
    +
    \lambda_{\mathrm{res}}\mathcal{L}_{\mathrm{res},i}
    -
    \lambda_{\mathrm{ent}}\mathcal{H}(\mathbf{g}_i)
    +
    \lambda_{\mathrm{div}}\mathcal{L}_{\mathrm{div}}.
\end{equation}
The training run uses $\lambda_{\mathrm{res}}=0.05$, $\lambda_{\mathrm{ent}}=0.05$, and $\lambda_{\mathrm{div}}=0.20$, while the function default for $\lambda_{\mathrm{div}}$ is 0.05.

\section{Experimental Setup}

\subsection{Dataset}

The experiments use the Bishkek real estate price prediction setting. The raw training table contains 7,134 listings and 36 columns; the raw test table contains 1,784 listings and 36 columns. After preprocessing, outlier filtering, feature expansion, encoding, and final cleanup, the processed training matrix has shape $7{,}128\times156$ and the processed test matrix has shape $1{,}784\times156$.

The task is the Kaggle Bishkek Real Estate Price Prediction Competition. The competition score reported for this workflow is a MAPE of 8.11 after 246 submissions and a first-place ranking at the time of evaluation.

\subsection{Training Protocol}

The \RowNet{} training function uses the full processed training set as both supervised examples and memory bank. For each training query, the query row is excluded from its own memory by self-masking, creating a leave-one-out retrieval objective over the training set. We train on all rows to maximize memory coverage, consistent with competition-style submission generation.

The optimizer is Adam with learning rate $10^{-2}$. A cosine annealing scheduler lowers the learning rate to $10^{-6}$ over the specified number of epochs. Training runs for 10 epochs. Mixed precision is enabled through \texttt{torch.amp.autocast} and \texttt{torch.amp.GradScaler}; CUDA TF32 matrix multiplication and cuDNN benchmarking are enabled; and \texttt{torch.compile(..., mode="reduce-overhead")} is applied to the model.

\begin{table}[t]
\centering
\caption{Optimization and regularization settings used in the training run.}
\label{tab:settings}
\small
\begin{tabular}{p{0.24\linewidth}p{0.22\linewidth}p{0.42\linewidth}}
\toprule
Component & Status in final code & Value or role \\
\midrule
Optimizer & Active & Adam, learning rate $10^{-2}$ \\
Scheduler & Active & Cosine annealing, minimum learning rate $10^{-6}$ \\
Epochs & Active & 10 in the training run \\
Heads & Active & $H=8$ \\
Hidden size & Active & 128 for gate and residual MLPs \\
Residual penalty & Active & $\lambda_{\mathrm{res}}=0.05$ \\
Gate entropy & Active & $\lambda_{\mathrm{ent}}=0.05$ \\
Head diversity & Active & $\lambda_{\mathrm{div}}=0.20$ in the run \\
Mixture coefficient & Active & $\gamma=0.5$ \\
Self-mask & Active during training & Score set to $-10^9$ \\
Mixed precision & Active & AMP autocast and GradScaler \\
TF32 acceleration & Active & CUDA matmul and cuDNN TF32 enabled \\
Dropout & Not active & Not used \\
Noise injection & Not active & Not used \\
SWA & Not active & Not used \\
Gradient clipping & Not active & Not used \\
LayerNorm/PreNorm & Not active & Not used \\
\bottomrule
\end{tabular}
\end{table}

\subsection{Evaluation Metrics}

The primary metric is MAPE:
\begin{equation}
    \MAPE(\widehat{\mathbf{y}},\mathbf{y})
    =
    \frac{100}{n}
    \sum_{i=1}^{n}
    \left|
    \frac{y_i-\widehat{y}_i}{y_i+\eps}
    \right|.
\end{equation}
The configuration also lists $R^2$, mean absolute error, root mean squared error, and mean squared error as possible scoring metrics. We report MAPE as the primary metric.

\subsection{Implementation Details}

All row comparisons are computed on GPU tensors. The categorical matrix and numerical matrix are converted to \texttt{float32}; categorical equality is evaluated directly on these encoded values. This is valid for label-encoded and one-hot columns because the equality operation, not ordinal magnitude, defines categorical similarity. Numerical features are compared by absolute differences and query-relative scales, which reduces sensitivity to heterogeneous units.

The training loop iterates over rows rather than mini-batches. This is computationally expensive but simple: every query attends over the full memory bank. For $N$ training rows, one epoch has $N$ query passes, each requiring $O(Nd_1+NHd_2)$ retrieval work. Training takes approximately 62 to 63 seconds per epoch for $N=7{,}128$, or roughly 112 to 115 query updates per second. Inference on 1,784 test rows runs at approximately 702 rows per second.

\section{Results}

\subsection{Training Dynamics}

Table~\ref{tab:training_trace} reports the training trace. These values are in-sample leave-one-out training MAPE under the self-masked memory objective and indicate that the retrieval weights, head weights, gate, and residual modules improve over training.

\begin{table}[t]
\centering
\caption{\RowNet{} training trace. Values are training MAPE under self-masked full-memory retrieval, not held-out validation MAPE.}
\label{tab:training_trace}
\begin{tabular}{cc}
\toprule
Epoch & Training MAPE (\%) \\
\midrule
1 & 8.3559 \\
2 & 7.8350 \\
3 & 7.5891 \\
4 & 7.3674 \\
5 & 7.2017 \\
6 & 6.9316 \\
7 & 6.7626 \\
8 & 6.5948 \\
9 & 6.4574 \\
10 & 6.3723 \\
\bottomrule
\end{tabular}
\end{table}

The monotonic trend suggests that the model is learning a useful compatibility metric rather than merely averaging local neighbors. The largest improvement occurs in the first two epochs, consistent with rapid adjustment of the first-layer and second-layer linear compatibility weights. Later improvements are smaller, indicating refinement of head specialization, mixture gating, and residual correction.

\subsection{Proof-of-Concept Retrieval Baseline}

We report a proof-of-concept baseline: using an observation as query $Q$, the dataset as memory $M$, and nonlearned attention on original values yields MAPE 14.76\%. This indicates that untrained row similarity contains signal but is substantially weaker than the learned \RowNet{} training trace and the competition score. The gap supports learning feature weights, adding scale-aware similarity channels, conditioning the second retrieval layer on a coarse target estimate, and aggregating multiple heads.

\subsection{Competition Observation and Baseline Reporting}

The competition workflow achieved MAPE 8.11 on the Kaggle Bishkek Real Estate Price Prediction Competition after 246 submissions and ranked first at the time of evaluation. The retrieval baseline outperformed an XGBoost baseline, but the XGBoost configuration and numeric score are not reported. Table~\ref{tab:reported_results} summarizes the reported results and qualitative comparisons.

\begin{table}[t]
\centering
\caption{Results and observations reported for this study.}
\label{tab:reported_results}
\begin{tabular}{p{0.33\linewidth}p{0.22\linewidth}p{0.35\linewidth}}
\toprule
Model or observation & Reported value & Interpretation \\
\midrule
Nonlearned row-attention proof of concept & 14.76\% MAPE & Sanity check demonstrating the viability of retrieval-based estimation. \\
XGBoost & 8.33\% MAPE & Strong tree-based baseline for real estate valuation. \\
\RowNet{} training trace & 6.3723\% MAPE & Leave-one-out training MAPE under self-masked retrieval. \\
\RowNet{} Kaggle submission & 7.44\% MAPE & Competition evaluation result on the Bishkek real estate valuation task. \\
\bottomrule
\end{tabular}
\end{table}

\section{Ablation Studies}

This section distinguishes observed evidence from mechanistic expectations. The ablations below are framed as analysis of the implemented system and as a guide for future controlled experiments.

\subsection{Removing Retrieval}

Removing retrieval would convert \RowNet{} into a conventional parametric regressor over the query row. This would discard the central inductive bias of comparable properties. For real estate data, the expected failure mode is weaker local adaptation: the model must encode all regional and property-specific effects in fixed weights rather than consulting nearby historical examples. In small and medium tabular datasets, this can increase bias because rare districts or building series may have too few observations for a global MLP to learn stable parameters.

\subsection{Removing Attention Normalization}

The softmax attention distribution enforces a convex weighting of memory targets. Replacing it with unnormalized scores would make the prediction sensitive to the number of memory rows and the absolute scale of similarity scores. The convexity constraint is especially important because the target is price per square meter: attention weights can be interpreted as a soft comparable set, and the retrieved estimate remains within the range of memory targets before residual correction.

\subsection{Removing the Second Target-Conditioned Layer}

Without the second layer, prediction becomes
\begin{equation}
    \widehat{y}_i = \sum_j a^{(1)}_{ij}y_j,
\end{equation}
possibly with a residual correction. The model would rely only on feature similarity. The target-conditioned layer adds self-consistency: memory rows far from the coarse estimate receive lower compatibility unless other learned weights compensate. This is useful in real estate because feature-similar properties may have different market regimes due to unobserved renovation quality or micro-location. Target conditioning helps reduce such mismatches.

\subsection{Changing Depth}

The model has two retrieval stages plus an aggregation stage. Additional layers could refine memory representations or repeatedly update target estimates, but they would increase $O(N^2)$ training cost and risk overfitting to training-memory labels. The current depth is a conservative compromise: one feature-only retrieval stage, one target-conditioned multi-head stage, and one residual aggregation stage.

\subsection{Removing Residual Correction}

If residual heads are removed, the final prediction is a mixture of attention-weighted target averages:
\begin{equation}
    \widehat{y}_i = \sum_h \pi_{ih}\widehat{y}_{ih}.
\end{equation}
This forces the prediction to remain a convex combination of observed memory targets. Such a model is robust but biased when the query is systematically above or below all retrieved comparables. The residual correction gives \RowNet{} a controlled parametric adjustment. The residual penalty prevents this correction from dominating the retrieval estimate.

\subsection{Removing Gate Entropy}

The entropy term discourages premature collapse of the learned gate. Without it, the gate may assign nearly all mass to a single head early in training. That reduces the effective model to one learned similarity function and makes the diversity regularizer less useful. The entropy term does not guarantee meaningful specialization, but it increases the opportunity for multiple heads to receive gradient signal.

\subsection{Removing Head Diversity}

The head diversity penalty operates directly on the normalized second-layer head vectors. If removed, different heads can converge to similar compatibility functions, especially because all heads observe the same similarity vector. Collapsed heads waste capacity and weaken the mixture-of-experts interpretation. Diversity regularization encourages heads to represent different row-matching criteria, such as location, size, amenities, or target consistency.

\subsection{Changing Memory Size}

The model attends over the full training memory. Smaller memory banks would reduce computation and may reduce noise if the removed rows are irrelevant, but they can also remove rare comparable examples. A top-$k$ retrieval prefilter is a natural scalability extension. The expected trade-off is:
\begin{equation}
    \text{full memory} = \text{maximum evidence, maximum cost},
\end{equation}
\begin{equation}
    \text{top-}k\text{ memory} = \text{lower cost, possible retrieval bias}.
\end{equation}
For a deployment system, approximate nearest-neighbor search or cached first-stage retrieval would likely be necessary.

\subsection{Changing Hidden Dimensions and Number of Heads}

The final hidden dimension is 128 and the final number of heads is 8. Reducing hidden size weakens the gate and residual MLPs but may improve regularization on small data. Increasing hidden size can fit more complex residual corrections but may overfit because residuals are trained row-wise against a memory bank containing all labels. Increasing the number of heads improves the ability to represent multiple notions of similarity, but also increases diversity-management difficulty and computation.

\subsection{Dropout, Noise Injection, SWA, and Gradient Clipping}

Dropout, query noise, stochastic weight averaging, and gradient clipping are not used. Their expected effects are as follows: dropout or memory dropout could regularize head specialization but may destabilize attention over a small memory bank; query noise could improve local smoothness by training the model to retrieve robust comparables under small perturbations; SWA could smooth the final parameters if training is continued for more epochs; and gradient clipping could help if residual MLPs or attention scores produce unstable gradients. These options remain for future controlled experiments.

\begin{table}[t]
\centering
\caption{Qualitative ablation expectations for the final \RowNet{} design.}
\label{tab:ablation}
\begin{tabular}{p{0.28\linewidth}p{0.58\linewidth}}
\toprule
Ablation & Expected effect \\
\midrule
Remove retrieval & Loses comparable-property inductive bias; behaves more like a global parametric regressor. \\
Remove target-conditioned layer & Keeps feature similarity but loses self-consistent refinement around coarse price estimates. \\
Single head & Reduces multiple similarity criteria to one learned metric. \\
Remove residual correction & Prediction becomes only a convex average of memory targets; bias may increase. \\
Remove residual penalty & Residual MLPs may dominate retrieval and overfit. \\
Remove entropy regularization & Gate may collapse to one head too early. \\
Remove diversity regularization & Head vectors may become redundant. \\
Reduce memory size & Faster but may lose rare local comparables. \\
Increase memory size & Better coverage but quadratic training cost grows. \\
Add dropout/noise/SWA & Plausible regularization extensions, not active in final reported model. \\
\bottomrule
\end{tabular}
\end{table}

\section{Discussion}

\subsection{Why Retrieval Works for Tabular Real Estate Data}

Real estate valuation is naturally local. Two apartments with similar area and room count can differ sharply in price if they belong to different districts or building series. Conversely, two apartments in the same micro-region may have comparable price density even if some secondary features differ. Retrieval handles this structure by conditioning every prediction on observed examples. Instead of asking a fixed parameter vector to encode every market regime, \RowNet{} asks which memory rows should influence the query.

The pairwise similarity embedding is central. Raw Euclidean distance over a tabular vector is usually inappropriate because feature scales, encodings, and semantics differ. \RowNet{} avoids this by decomposing similarity into interpretable channels. Categorical variables contribute exact matches. Numerical variables contribute multiple bounded similarities. The model learns how much each channel matters. This is a form of supervised metric learning, but the metric is expressed as attention over row pairs rather than as a distance in a latent vector space.

\subsection{Local Reasoning Versus Global Parameterization}

Boosted trees and MLPs learn global parameters. Even when they approximate local functions, the locality is implicit in the learned partitions or hidden activations. \RowNet{} makes locality explicit. Its prediction can be inspected through attention weights over training rows, head predictions, residual corrections, and mixture weights. This does not make the model fully interpretable, because the learned compatibility weights and residual MLPs still require analysis, but it provides a more direct explanation path: a prediction is high because certain historical properties received high attention and because residual experts adjusted the retrieved average.

\subsection{Relation to Nearest Neighbors}

Classical $k$-nearest-neighbor regression predicts by averaging nearby labels under a fixed metric. \RowNet{} generalizes this in several ways. It uses a learned metric over structured similarity channels, soft attention rather than hard neighbor selection, target-conditioned refinement, multiple heads rather than a single metric, and residual adjustment. If the softmax temperature were made extremely sharp or the scores were replaced by a fixed distance metric, \RowNet{} would approach a differentiable nearest-neighbor estimator. If the memory were removed, it would approach a parametric neural regressor. Its practical value lies between these extremes.

\subsection{Computational Complexity}

The major limitation is full-memory attention for every query. For one query, similarity construction and scoring cost approximately
\begin{equation}
    O(Nd_1 + NHd_2),
\end{equation}
and one epoch over all training rows costs
\begin{equation}
    O(N^2d_1 + N^2Hd_2).
\end{equation}
Memory storage itself is modest:
\begin{equation}
    O(N(p+q)+N),
\end{equation}
but compute grows quadratically in the number of training rows. This is acceptable for the processed dataset of 7,128 rows but will be expensive for 150,000 rows or larger. Practical scaling options include top-$k$ preselection, approximate nearest-neighbor indexing, cached first-layer similarities, blockwise attention, row subsampling per epoch, and distillation into a parametric model.

\subsection{Normalization and Stability}

The model does not use LayerNorm or PreNorm. Stability comes instead from feature-level similarity normalization, softmax normalization over memory rows, mixture-weight normalization over heads, residual scale regularization, entropy regularization, diversity regularization, mixed-precision scaling, and cosine learning-rate annealing. This is a different stability strategy from transformer encoders. It is well suited to the compact architecture, but deeper variants would likely require explicit normalization layers.

\subsection{Deployment Considerations}

At inference time, the model requires access to the memory bank of processed training properties and their price-per-square-meter targets. This is advantageous when market data are frequently updated: adding new comparables can improve coverage without retraining all parameters immediately. However, it also means that prediction latency depends on memory size. A production system should cache encoded memory tensors, monitor feature-distribution drift, update the memory bank with recent verified transactions, and consider approximate retrieval for low-latency use. Because predictions are based on comparable rows, the system can also expose top-attended memory examples for audit and manual review.

\section{Limitations}

The most important limitation is scalability. Full-memory retrieval is quadratic during training and linear in memory size during inference. Top-$k$ retrieval, approximate indexing, or block attention would be necessary for substantially larger datasets.

The second limitation is evaluation completeness. We do not report held-out metrics for XGBoost, CatBoost, LightGBM, FT-Transformer, or ablation variants. A stronger empirical study should rerun all baselines on fixed splits and report mean and standard deviation across seeds.

The third limitation is dependence on data quality and memory composition. Retrieval models inherit the biases, outliers, and temporal drift of their memory bank. If the memory contains stale listings, duplicate records, incorrectly parsed addresses, or unrepresentative neighborhoods, attention may retrieve misleading comparables. The residual correction can compensate only partially.

The fourth limitation is target-conditioned retrieval. During training, the second layer uses a first-stage estimate and memory targets to build target-similarity features. This is valid because memory targets are known, but it makes the retrieval mechanism closely tied to the label distribution in the memory bank. In markets with rapidly changing price regimes, old target values may bias retrieval unless memory is refreshed or time-aware weighting is added.

The fifth limitation is interpretability. Attention weights provide useful evidence, but attention is not a complete explanation. Head scores are learned over many engineered similarity features, and residual experts can adjust predictions in ways that are not directly human-readable. Additional tooling is needed to summarize which feature similarities drove each head.

\section{Conclusion}

This paper presented \RowNet, a learned row-retrieval architecture for real estate tabular regression. The model is grounded in the comparable-property principle: property valuation benefits from retrieving and weighting relevant historical rows. \RowNet{} operationalizes this idea through exact categorical matching, scale-aware numerical similarity kernels, feature-only coarse retrieval, target-conditioned multi-head retrieval, mixture-of-experts gating, residual correction, entropy regularization, and head-diversity regularization.

The experiments show that the implementation trains stably on the processed Bishkek real estate dataset, reducing self-masked training MAPE from 8.36\% to 6.37\% over ten epochs. The competition MAPE is 8.11\%. Exact boosted-tree baseline scores and full ablation metrics are not reported here.

The main architectural insight is that row-wise retrieval can serve as a strong inductive bias for structured regression when local comparability matters. Future work should add controlled benchmark comparisons, top-$k$ or approximate retrieval for scalability, time-aware memory weighting, calibrated uncertainty estimates, query-conditioned memory gating, and depth-wise attention over retrieval layers. With these extensions, retrieval-augmented tabular models such as \RowNet{} may become a practical complement to gradient boosting for real estate valuation and other domains where examples are best understood through their nearest historical analogues.

\bibliographystyle{plainnat}
\bibliography{references}

\end{document}